# Spiral Model Technique For Data Science & Machine Learning Lifecycle

*Rohith Mahadevan, mailrohithmahadevan@gmail.com*

**Abstract**

Analytics play an important role in modern business. Companies adapt data science lifecycles to their culture to seek productivity and improve their competitiveness among others. Data science lifecycles are fairly an important contributing factor to start and end a project that are data dependent. Data science life cycle comprises of series of steps that are involved in a project. A typical life cycle states that it is a linear or cyclical model that revolves around. It is mostly depicted that it is possible in a traditional data science life cycle to start the process again after reaching the end of cycle. This paper suggests a new technique to incorporate data science life cycle to business problems that have a clear end goal. A new technique called spiral technique is introduced to emphasize versatility, agility and iterative approach to business processes.

**Introduction:**

Data Science problems have become increasingly popular in modern times. An iterative approach to solving business data problems is common with data science applications. Data Science life cycle is defined as a series of steps that is necessary to deliver the project or to conduct an analysis. Every data science project is different and sometimes ambiguous. However, most of the projects tend to flow through some series of generalized steps that are called data science life cycle steps [1]. The business world relies more on data science projects nowadays. Data science has become an important contributing factor to our society, and we have become reliant on this technology for more opportunities. Dinh-An Ho (2020) discussed about how important data science technology is used and what are the risks involved in delivering projects using data science life.

A standard framework for delivering data science projects is CRISP-DM (Cross Industry Standard Process for Data Mining. Vijay et al [3] in their paper discussed utilizing the Cross Industry Standard Process for Data Mining (CRISP-DM) framework as a basis for the data science process. This framework provides a structured approach to solving data science problems and is widely adopted in the industry. The paper discusses the process of gathering subject matter expertise, exploring data through statistics and visualization, building models using data science algorithms, testing these models, and deploying them in a production environment. This iterative process aims to uncover patterns effectively and solve data science problems efficiently. All these frameworks depend on the data that we process for the use case. Different problems have different sources of data. Data is a valuable entity for any business use case. Data Management involves the practice of collecting, analyzing and using data securely, efficiently and cost-effectively for the purpose of the business problem. Kumar et al (2020) in their paper discussed about using data management and the benefits of using data science life cycle in healthcare industry. The paper also discussed the challenges and how to improve them based on the cycle [4].

Data science life cycle comprises of series of steps involved from cleaning to deploying based on the business use case. Each step is responsible for delivering an issue about the data that we are dealing with. It is necessary to pay attention to such steps as it is critical. The paper by Haertel et al (2022) talks about the documentation artifacts that are necessary for ensuring traceability, reproducibility, and knowledge retention across the data science life cycle, particularly emphasizing the importance of comprehensive documentation for improving project management practices [5]. The relationship between the data and the machine learning model that we build should have high correlation in each case. This correlation can be achieved only through a successful series of data science steps. Each step is responsible for identifying the Key performance indicators (KPIs) that contribute towards the success of the model. Yadav et al. (2022) discussed how organizations can utilize data science methodology to make crucial business decisions by understanding its principles [6]. The first step that indicates the success of the model is understanding the business problem statement. Lohmann (2021) provides insights into the foundational structures and concepts of business intelligence and data science, contextualizing them within modern digitalized management [7]. Once business understanding is clear, the next step would involve data collection.

**Steps involved in data science life cycle**

**Data Collection**

Data collection involves collecting and aggregating data from different data sources.

**Data Preparation**

Data preparation is a critical step that involves transforming raw data into a format suitable for analysis [8]. This step includes data cleaning, integration, transformation, and reduction. Common challenges include handling missing values, dealing with outliers, and ensuring data consistency. Solutions involve using imputation techniques, normalization, and aggregation methods.

**Data Cleaning and Preprocessing**

Data cleaning involves removing or correcting inaccuracies and inconsistencies in the data. Preprocessing includes normalization, encoding categorical variables, and scaling features. Some of the popular techniques like outlier detection, handling missing values, and data transformation can be used [9].

**Data Exploration and Visualization**

Data exploration helps in understanding the data's underlying patterns and relationships. Visualization aids in communicating these insights effectively [10].

**Feature Engineering**

Feature engineering is the process of creating new features or modifying existing ones to improve model performance [11]. This includes feature selection, extraction, and creation. Feature selection techniques like recursive feature elimination (RFE) and principal component analysis (PCA) are commonly used.

**Model Building and evaluation:**

Algorithms and Approaches: Various machine learning algorithms, including regression, classification, clustering, and ensemble methods, are used depending on the problem [12].

Model evaluation ensures that the model performs well on unseen data. Validation techniques like cross-validation and bootstrapping are essential. Metrics such as accuracy, precision, recall, F1 score, and ROC-AUC are used to evaluate model performance.

**Model Deployment and maintenance:**

Model deployment involves deploying the model in real time for the required use case. This can be integrated with any back-end application or to a web server. Monitoring and maintenance ensure the model continues to perform well over time. It involves tracking performance metrics and updating the model as needed [13].

**Spiral Technique:**

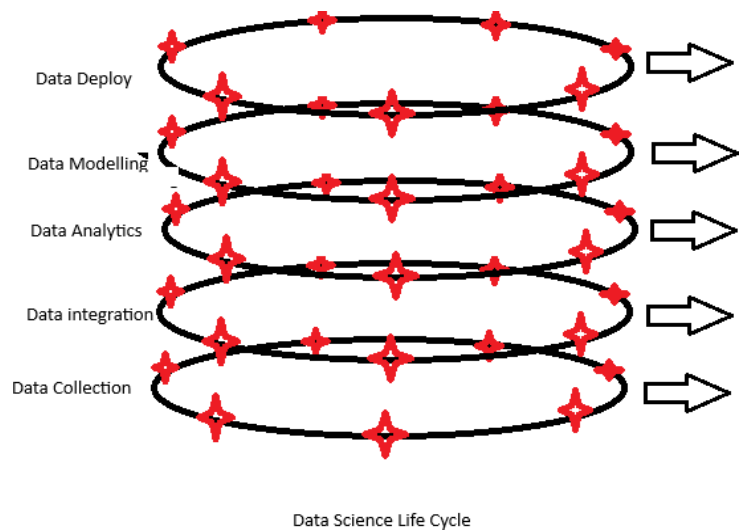

Data Science Life Cycle

The spiral technique involves a series of cycles comprised together to form a spiral that has exits in all directions. Each revolution involves a milestone in data science life cycle. At the end of each revolution there are flags that need to be set. These flags can be used to determine whether we need to traverse back or move forward to the next step. If the requirements are satisfied, the flag also should be an indicator to have an exit from the life cycle.

2\. Data integration:

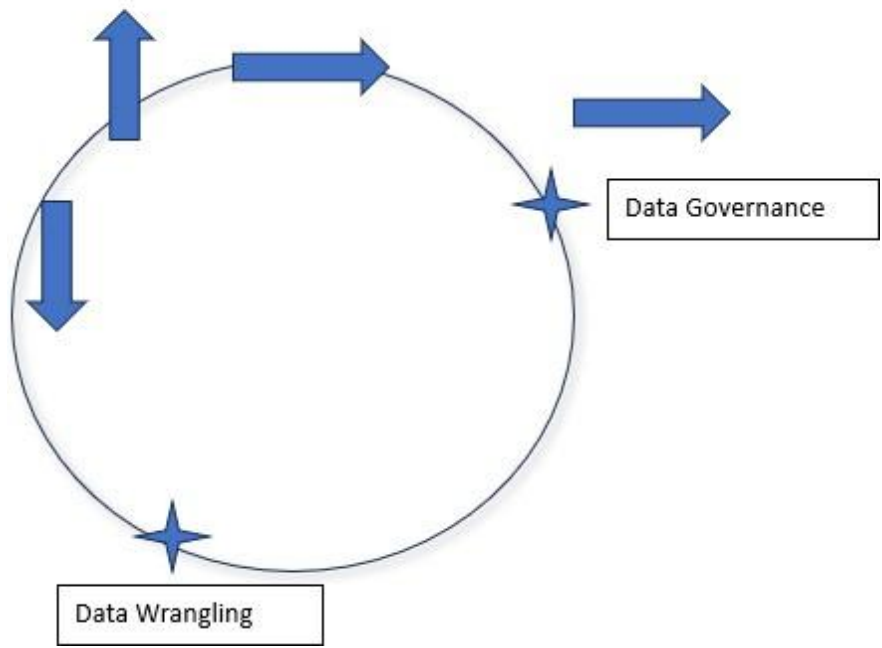

**Purpose of the spiral technique:**

The main purpose of the spiral technique is to understand efficiently about exiting the data science life cycle at the appropriate time to save resources. However, the exit can be a challenge depending on the use case of the model. In the previous discussion, the life cycle for a complete product is developed using the cycle. The cycle is synchronized with the linear progress of the process. Exiting can be anywhere depending on the cycle. The COVID case study discusses creating a unified dataset for the process. This seems like an end- to-end cycle. However, the data is stacked each week and updated accordingly. If we use a traditional model for this approach, it simply portraits that we exit once the step is completed. Data Science life cycle is a versatile model that has its own sub cycles. Each sub cycle can create a process and paves a way to exit.

## A. Case Study 1:

**Covid Life cycle:**

The covid cycle is a great example of how the model should be displayed. For our understanding, let us keep a time life of 5 weeks. The spiral model is only applicable when the business problem is well defined. Even if the exit is not known in the initial days, it is possible to create or assume an exit as we progress in the cycle. The business problem or end goal is 'to create a unified dataset'. The dataset is not static and dynamically changed as week progresses. In the initial phase, the data collection is started and data from different sources [as discussed in the research paper] is processed. This is our second stage of the cycle [The first step is identifying the business problem; in this case the end goal is clearly mentioned. The second step is data collection]. Once the data is collected the paper does not talk about how this data will be utilized. Instead, it talks about the granularity of the data level. The third step is consolidating the data and pushing the data to a common source. For our assumption let us say that the data is stored in some server and produced as a CSV file. The typical cycle would end here. If the data is static, then it is evident that the cycle might end here. However, it is not static data. The next set of data comes next week, and it's updated. The traditional approach of any data life cycle suggests that all these steps are repeated but only the new and fresh data is incremented. The incremental data is then merged with the existing data and pushed as a new dataset.

**i) Modified approach for the data life cycle:**

The modified approach would be to describe a flag at the end of each week. This flag stands a checkpoint indicating the incremental data refresh. The checkpoint is then taken at the end of each week and then the cycle continues. Therefore, for our assumption we created a 5-week window.

Number of weeks – 5

Number of iterations – 5

True exit – 1

Periodical exit – 4

Number of flags – (iterations – True exit) = 5 – 1 = 4

The cycle would spin as a spiral and not as a circle indicating a clear exit. Here the exit is instantaneous at the end of each spin. This creates a unified data model at the end of each spin and helps us to be in check of the life cycle.

**ii) Advantages of using the life cycle:**

As a clear exit is mentioned, the project can be well defined and well established. The exit helps us to be more productive and not to concentrate on unwanted steps like data analysis or establishing a model with covid data. This is a great example of how the cycle is used for a minimalistic approach. Flags indicate the clear outcome of the model.

In a traditional lifecycle, each week's new data would be treated as a separate cycle but without explicit stopping criteria – teams might keep preparing data indefinitely or proceed to analysis unnecessarily. In our spiral model, we knew to stop at week 5 once the unified dataset was complete, and we intentionally did not perform unrelated steps like predictive modeling.

**B. Case Study 2:**

Predicting Employee Turnover + Designing Retention Policies

**Goal & Exit Criterion (Business Understanding)**

Objective: Build a model to predict which employees are likely to churn (leave) within next 6 months, with target performance metrics (e.g., ROC AUC ≥ 0.85, precision/recall thresholds).

True Exit Flag Criterion: Once the predictive model meets the business threshold and the retention policy design yields expected uplift, the spiral can exit.

**Iteration / Revolution 1**

Data Collection: Gather HR data: demographics, performance, attendance, compensation, tenure, job role, etc. Use the same data sources that Ribes et al. used (they aggregated from HR systems).

Data Preparation: Clean missing values, encode categorical features, normalize continuous variables.

Feature Engineering: Start with basic features (age, years at company, salary), then evaluate their predictive power.

Model Training / Evaluation: Train baseline models (logistic regression, random forest). Evaluate via cross-validation. Suppose ROC AUC = 0.76.

Check Flag: Performance is below threshold (0.85). So decision: continue.

**Iteration / Revolution 2**

Collect Additional Features: From HR surveys, engagement scores, promotion history, manager feedback.

Feature Engineering: Add interaction features, temporal trends (e.g. change in performance over time).

Model Retraining: Re-run with more features. Suppose model now gives AUC = 0.82.

Check Flag: Still below threshold. Continue.

**Iteration / Revolution 3**

Refinement & Policy Modeling: Add features like external market benchmarks or external job openings in region.

Model Training: Advanced models (boosting, ensemble). Achieve AUC = 0.87.

Design Retention Policy: Using model predictions, simulate retention interventions (e.g. offering promotion, bonus, role change) and estimate cost/benefit.

Check Flag: Both predictive performance and retention policy simulation meet desired business goals → True Exit.

**Deployment & Monitoring**

- Deploy the model to HR system for periodic scoring (e.g. monthly).

- Monitor model drift (if features shift) or drop in performance.

- Use periodic exits only when retraining is needed (if performance degrades) and re-run spiral loops then.

**Discussion**

The predictive model case illustrates how the Spiral Model Technique enhances both agility and accountability in machine learning workflows. By embedding exit checkpoints within each revolution, the lifecycle adapts dynamically — allowing data scientists to stop iterations when business-defined thresholds are achieved.

This mirrors concepts discussed by Amershi et al. (2019) on Software Engineering for Machine Learning (Communications of the ACM, 62(9): 62–70) and Sculley et al. (2015) on Hidden Technical Debt in Machine Learning Systems (NIPS), both of which emphasize the risk of unbounded iteration and maintenance costs in ML pipelines.

In essence, the Spiral Model transforms the ML lifecycle into a goal-driven, resource-efficient framework where teams move purposefully, exit deliberately, and iterate only when meaningful.

**Acknowledgement:**

I thank my professor Dr. Christopher Teploz for encouraging me and mentoring me to write this paper. His valuable advice and guidance are much appreciated and contributed towards the success of this paper.